# Mechanical design of a distal scanner for confocal microlaparoscope: a conic solution*

Mustafa Suphi Erden, *Member, IEEE*, Benoît Rosa, *Student Member, IEEE*, Jérôme Szewczyk, and Guillaume Morel, *Member, IEEE*

*Abstract*—This paper presents the mechanical design of a distal scanner to perform a spiral scan for mosaic-imaging with a confocal microlaparoscope. First, it is demonstrated with *ex vivo* experiments that a spiral scan performs better than a raster scan on soft tissue. Then a mechanical design is developed in order to perform the spiral scan. The design in this paper is based on a conic structure with a particular curved surface. The mechanism is simple to implement and to drive; therefore, it is a low-cost solution. A 5:1 scale prototype is implemented by rapid prototyping and the requirements are validated by experiments. The experiments include manual and motor drive of the system. The manual drive demonstrates the resulting spiral motion by drawing the tip trajectory with an attached pencil. The motor drive demonstrates the speed control of the system with an analysis of video thread capturing the trajectory of a laser beam emitted from the tip.

## I. INTRODUCTION

THIS paper presents the mechanical design of a scanning system for mosaic-imaging with a confocal microlaparoscope. Confocal microlaparoscope imaging is a promising approach for detecting cancer cells on the outside tissues of organs, such as in the abdominal cavity [1]. In many cases it might replace the invasive and time consuming conventional biopsy, in which tissue samples are taken out of the body and examined under microscope.

There have been a few different designs of confocal micro-imaging of living tissues for microlaparoscopy [2] and closely related microendoscopy [3, 4, 5, 6]. The images obtained by such micro-imaging typically cover an area of $240 \times 200$ $\mu m^2$. The image size is small; this is because of the necessity of fine resolution and because the optic lenses and the fiber cable are minimized for minimal invasiveness. An image of such size is not enough for a conclusive diagnosis, which typically requires an image size of around 3 $mm^2$.

A solution to obtain such large scale images is to scan the region of interest and merge the images by mosaicing algorithms [7, 8]. The study [9] demonstrates the applicability of this approach *in vivo* on human patients by manually passing the miniprobe over the region of interest and using the mosaicing algorithm in [10].

The design presented in this paper is for imaging the outside tissues of organs in the abdominal cavity for detecting cancer cells. We use the correlation-function-based mosaicing algorithm presented in [10, 11], designed for *in vivo* tissue imaging. For our purposes and with this mosaicing algorithm, it requires to maintain a distance of 150 μm from a previously traced line with an approximate precision of 25 μm for a continuous duration of approximately one minute. With manual sweeping it is difficult to maintain such precision. An assistive handheld instrument is presented for micro-positioning for intraocular laser surgery [12], but it is large to be used in minimally invasive surgery and not intended for one minute long continuous manipulation. The motorized surgical microscope presented in [13] enables the surgeon to control the movement of the microscope by index finger movements with a remote controller. Although this is a semi-automated system, it would be tedious for a surgeon to make the probe follow a proper scan path by finger movements.

In a study on automated scanning the authors perform image-mosaicing on human hand skin using a MEMS based design [14] and the mosaicing algorithm in [15]. The MEMS scanner is batch fabricated on silicon wafers with four deep-reactive-ion-etching steps. In this paper we present a lower-cost solution for the scanning problem.

In a recent study the latter three authors and colleagues demonstrate a design for micro-image scanning based on balloon catheters actuation [16]. The drawback of this system is that an active control is required for the movement of the probe that collects the images. The pneumatic control of the balloon catheters is not precise enough for a proper scan. In this paper, we develop a scanning system which is more robust and which uses a simple motor actuation.

The study [17] presents a wide-angle-view-endoscope which uses two articulated wedge prisms that can be rotated by two motors. The prisms that are noted to be under construction are 12 mm in diameter. We aim at having a mechanical system that can fit to a 5 mm inner diameter tube. Furthermore, we propose a solution which requires only one-degree-of-freedom rotation that can be provided by distal actuation.

The solution in this paper is based on using a conic structure and scanning the region by following an Archimedean spiral. With such a spiral the radius increases proportionally with the angle of turn. We present our research results that demonstrate the superiority of a spiral scan over a raster scan on soft tissue. We explain the conceptual principle of generating a spiral motion by using the conic structure and demonstrate the overall SolidWorks design. We present an enlarged prototype of the design to 5:1 scale dimensions produced by rapid prototyping and validate the design requirements by driving this 5:1 scale prototype both manually and with a motor.

* Research funded by OSEO (Maisons-Alfort, France) under the ISI Project PERSEE (number I0911038W).

M.S. Erden was with the Institut des Systèmes Intelligents et de Robotique (ISIR) at Université Pierre et Marie Curie (UPMC), Paris, France. He is now with the LASA Laboratory at the Engineering School of École Polytechnique Fédérale de Lausanne, CH 1015, Switzerland. (corresponding author: + 33 (0) 1 44 27 62 64 ; fax: +33.1.44.27.51.45; e-mail: mustafasuuphi.erden@gmail.com).

B. Rosa, J. Szewczyk, and G. Morel are with ISIR at UPMC, Paris, France (e-mail: {rosa, sz, morel}@isir.upmc.fr).

## II. REQUIREMENTS OF IN VIVO SCANNING

The previous study of micro-image scanning based on balloon catheters actuation [16] and our *ex vivo* experiments with a Stäubli-Robot (Stäubli-TX40) (Fig. 1) on soft tissue constitute the partial source of the requirements of our design. In this section we explain these requirements.

Our *ex vivo* imaging experiments presented in the next section make use of the Cellvizio imaging technology from Mauna Kea Technologies (Paris, France) [18]. The design introduced in this paper is also to be integrated with this system. The Cellvizio imaging technology performs confocal fluorescence imaging, records images in size 240×200 μm$^2$ with 4 μm lateral and 10 μm axial resolution at a rate of 12 frames/sec. The system is equipped with the mosaicing algorithm presented in [10, 18]. It can construct a mosaic out of sequentially collected set of images. The confocal probe of the system consists of a flexible bundle of optic fibers and an optic head hosting the micro-lenses. The outer diameter of the flexible bundle is 1.4 mm. The optic head is a 12 mm long cylinder with an outer diameter 2.6 mm. In this paper we refer to the optic head by "probe".

Our scanner is to be mounted in a tube 5 cm long with an inner diameter 5 mm. The surgeon manually brings the tip of the tube to the locus of interest and slightly presses on the tissue. The outer edge of the tube stabilizes the region to be scanned. The part of the tissue covered by the edges of the tube slightly bends and touches the tip of the probe inside. The probe is to be automatically swept on the tissue by the scanning system.

Our design is intended for automatic scan of an area 3 mm$^2$ on the soft tissue. This corresponds to a circular region approximately 1 mm in radius. Proper image mosaics are obtained when the scan speed is less than 0.5 mm/sec. Above 0.5 mm/sec, the deformation is increased with the increase in speed. The surgeons, with whom we collaborate, state that the duration of scan should be less than 3 minutes. In this paper we aim for 1 minute duration for a full spiral scan. The duration of 1 minute corresponds to a constant translation speed of approximately 0.38 mm/sec on spiral in unity scale dimensions. The distance between the two successive scan-lines of the spiral is restricted by the image size and should be less than 0.2 mm (200 μm); we aim for 0.15 mm in our design, in order to have overlap between the scan lines.

During the scan the tube is pressed on the tissue. The tissue and the probe get into contact inside the tube. This contact should be maintained throughout the scan while the mechanism is moving. The *ex vivo* experiments on beef liver show that the distance of the tip of the probe from a hypothetical flat surface touching the outer edge of the tube should be maintained between 200 and 300 μm. Distances outside this range avoid a proper contact between the tissue and the probe. We aim at maintaining a nominal distance of 200 μm. With our conic structure, the radius of scan is changed by inclination of the probe. This means that the inclination of the probe should not result in more than 100 μm (0.1 mm) change in tip point distance. The experiments also indicate that such inclination should not exceed 10 degrees for an acceptable imaging; best performance is achieved below 5 degrees.

## III. RASTER VERSUS SPIRAL SCAN

In this section we demonstrate our results of soft tissue scan that compare an Archimedean spiral path versus a raster path. We name the former as a *spiral scan* and the latter as a *raster scan*. The *raster scan* is noted in [8] as a readily performed solution; the spiral scan is mentioned to be better for mosaicing algorithms that use correlation of images.

The *ex vivo* experiments are performed with the aforementioned Stäubli-Robot on beef liver purchased from the supermarket (Fig. 1). We maintain an indentation depth 350±50 μm. We first bring the probe into contact with tissue in steps of 100 μm translation and then further translate a distance 300 μm. We detect the contact by observing the images on the monitor. In [19], it is demonstrated that the force vertically applied on soft tissue surface is proportional to the indentation depth; therefore we can assume that we maintain more or less the same vertical pressure in the *ex vivo* scans.

The robot is commanded to follow the raster and spiral trajectories shown by the dashed (violet) curves in Fig. 2(d) and Fig. 3(d). These are named as the *commanded trajectory*. The probe follows the light-solid (green) curves in the same figures with respect to the global reference frame. We name these as the *probe trajectory*. It is observed in these figures that the *probe trajectories* closely follow the *commanded trajectories*. The deviation from the *commanded trajectory* is largest when the robot makes large accelerations, at the corners of the raster scan and in the central region of the spiral scan. However, the deviation is less than 50 μm, which is one fourth of the smaller dimension of the field of view. Therefore the robot is precise enough to compare the raster and spiral scans.

The trajectory of the probe with respect to the tissue surface is named as the *image trajectory*. The *image trajectory* deviates from the *probe trajectory*; because the tissue also moves under the impact of the movement of the probe. The authors explain elsewhere the nature of the slip/stick phenomenon in soft tissue scan [20]. The *image trajectory* can be captured by analysis of the collected images. Actually the mosaicing algorithm [10] already performs this and returns the position of sequential images as they are placed in the resulting mosaic image. The dark-solid (black) lines in Fig. 2(d) and Fig. 3(d) show the *image trajectories* for the raster and spiral scans, respectively.

The question we would like the answer is which of the *raster* and *spiral* scans is better in the sense that the overall mosaic is compact and focused on the target point. An inspection of Fig. 2(d) reveals that with the raster scan, the *image trajectory* cannot follow the sharp turns (at the corners) of the *probe trajectory*. At these sharp turns there is no movement on the *image trajectory*. In Fig. 2 (e) and (f) these are observed as the instances when the *x* and *y* speeds of the *image trajectory* cannot catch up with the sudden rise of the speed of the *probe trajectory*. This means that the tissue moves with the probe. Therefore, the tissue is pushed in the direction of the probe and loaded with stress. The stress is released throughout the straight paths of the *probe trajectory* in the *x* direction. Consequently the segments of the *image trajectory* corresponding to these straight parts are inclined. These segments are getting shorter in time. This is probably because the successive movements along the short edges and in the same direction continuously increase the strain and change the friction force between the probe and tissue. The resulting mosaic lacks significant amount of the intended region on the left and right sides. The intended focus point is not at the center of the mosaic.

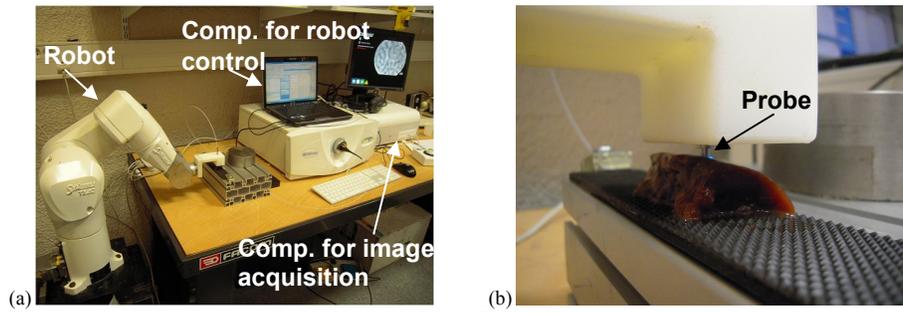

Fig. 1. Experimental setup for soft tissue scan. (a) Left to right: the Stäubli-Robot, the computer controlling the robot, and the computer performing the image acquisition. (b) The soft tissue (beef liver) under the probe attached to the end effector of the robot.

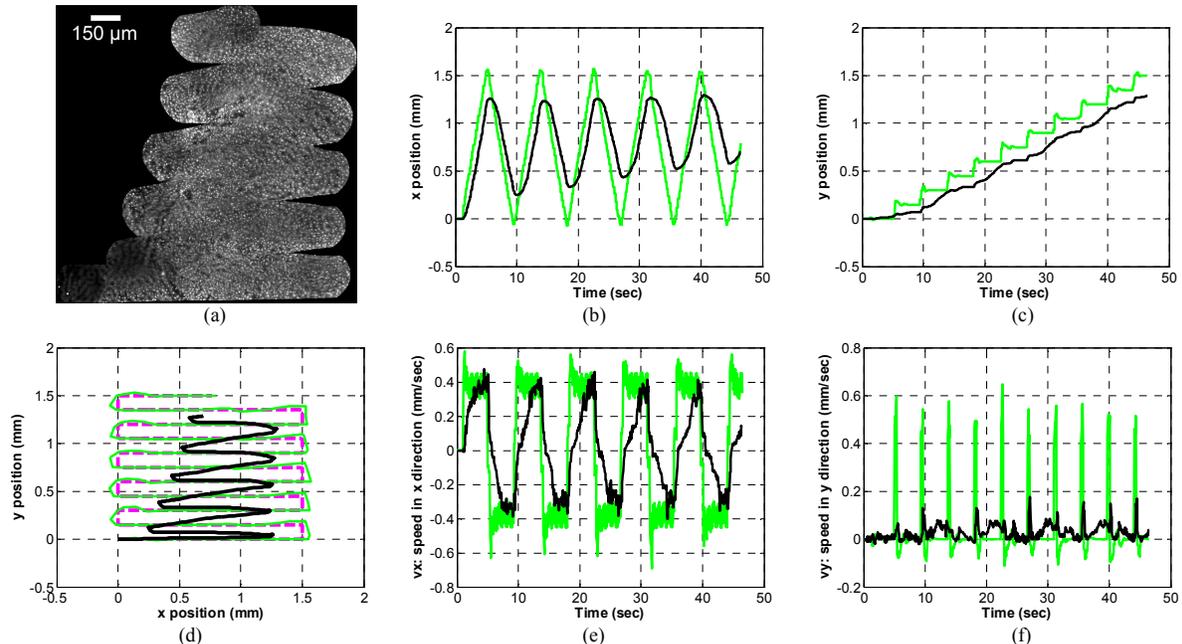

Fig. 2. Sample *raster scan* on *ex vivo* beef liver. (a) Mosaic image; (b) *x* position; (c) *y* position; (d) *x* versus *y* positions; (e) speed in *x* direction; (f) speed in *y* direction. Dark-solid (black) lines: *image trajectory*; light-solid (green): *probe trajectory*; light-dashed (violet): *commanded trajectory*.

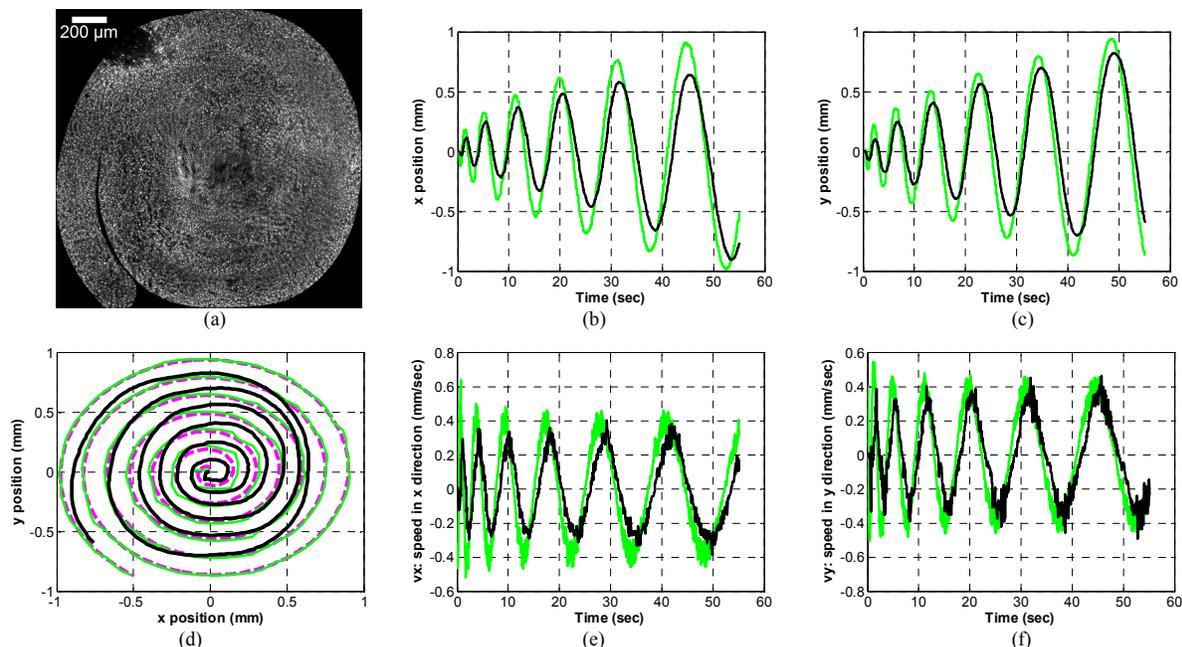

Fig. 3. Sample *spiral scan* on *ex vivo* beef liver. (a) Mosaic image; (b) *x* position; (c) *y* position; (d) *x* versus *y* positions; (e) speed in *x* direction; (f) speed in *y* direction. Dark-solid (black) lines: *image trajectory*; light-solid (green): *probe trajectory*; light-dashed (violet): *commanded trajectory*.

The spiral scan shown in Fig. 3(d) is free of sharp turns. The trajectories given in Fig. 3(e) and (f) reveal that there is no sudden rise of speed. Therefore, the loading and releasing of the stress on tissue is continuous and smooth. The speed corresponding to the *probe trajectory* and the *image trajectory* closely match. The resulting mosaic is compact. The focus point remains almost at the center of the image, well surrounded with a fully connected view.

In order to quantify these observations we perform calculations to demonstrate the mismatch between the *probe trajectories* and *image trajectories* for raster and spiral scans. We test both the position and velocity trajectories as a measure of mismatch with the equations (1) and (2), respectively. The mismatch values for the raster and spiral scans are indicated below the equations with the parameters *D* and *C*.

$$D = \int_0^{t_f} \frac{|p_i(t) - p_p(t)|}{t_f} dt \quad (1)$$
$$D_{raster} = 0.2363 \ mm$$
$$D_{spiral} = 0.2398 \ mm$$

$$C = \int_0^{t_f} \frac{|v_i(t) - v_p(t)|}{t_f} dt \quad (2)$$
$$C_{raster} = 0.2321 \ mm/sec$$
$$C_{spiral} = 0.1218 \ mm/sec$$

In (1) and (2), $p_i$ and $v_i$ are the position and velocity corresponding to the *image trajectory*, $p_p$ and $v_p$ are the ones corresponding to the *probe trajectory*, and $t_f$ is the final scan time. The measure *D* does not distinguish between the two scans. This is because the *image trajectory* is displaced with respect to the *probe trajectory* in both raster and spiral scans. The measure *C* clearly distinguishes between the raster and spiral scans. The velocity information is about the changes in direction of movement regardless of the position; therefore it better reveals the resemblance in shape even though the trajectories are displaced. According to the measure *C*, the mismatch between the speeds corresponding to the *probe trajectory* and the *image trajectory* is almost double for the raster scan in comparison to that of the spiral scan.

IV. CONCEPTUAL DESIGN OF THE CONIC SOLUTION

The requirement for the scanning tool is to generate a spiral motion at the tip point where the probe is located. The translation of the probe following an Archimedean spiral can be maintained by bringing together two coordinated motions: a rotation around the focus point and a change of the radius in proportion to the angle of rotation. It should be noted that the probe will only translate through the Archimedean spiral without any rotation around its axis. In our survey, we did not come across any mechanical solution that transfers simple rotation into a translation following an Archimedean spiral. The solution proposed here is using a conic structure as shown in Fig. 4. The conic part of this structure has an inclined surface (rightmost figure). The curve obtained by cross sectional cut of the cone can be represented by the relation between the parameters *f* and *s*.

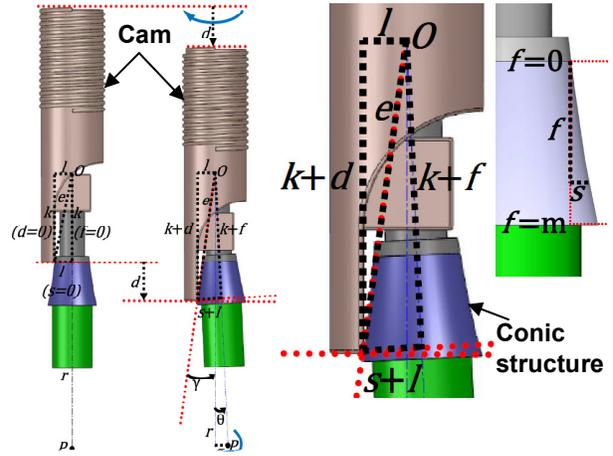

Fig. 4. Generation of the spiral motion at the tip point *P* with a cam pushing on the conic structure with the parameters used in the design of the conic surface.

Fig. 4 illustrates how this conic structure generates the spiral motion guided by a cam pushing and rotating on the conic surface. In these figures the conic structure is fixed to the point *O*; it can move around this point in two directions (*pitch* and *yaw*) but cannot turn around itself (no *roll* motion). This configuration is based on the assumption that for small angles of inclination ($\theta < 10^0$) a stiff cable will bend exactly at the point where it is fixed and the other parts of the cable will remain linear. It is also assumed that the conic structure is pressed on the cam that there is always contact between the comic surface and the tip of the cam. The fiber cable passes through the centerlines of the cam and conic structure.

In the left-hand side of Fig. 4 the guiding cam is at its nominal position (*d=0*); the center-lines of the cam and the conic structure coincide. In this situation the tip point *P* is located at the center of the region to be scanned. In the left-middle figure the guiding cam is translated (*d>0*). This causes the conic structure to incline with an angle $\theta$; as a result, the point *P* is translated by a distance $z = r \sin(\theta)$. The right-middle figure zooms the part of the conic structure where bending occurs. The amount of the translation of the cam (*d*) determines the amount of change in the tip position (*z*). The tip of the cam pushes the conic surface and translates the tip point *P* along a circle with radius *z*. With changing the iteration of the cam (*d*), the radius of this circle changes. In this way a spiral motion is obtained.

The cam should be rotated around its axis and translated in the direction of its tip. This can be achieved by a screw system between the rear part of the cam and the outer tube as shown in Fig. 4 (the tube is not shown). The translation is then proportional to the angle of rotation.

TABLE I
VALUES OF THE PARAMETERS IN THE FORMULATION AND SOLUTION

| Parameters | Value | Unit |
|---|---|---|
| $l$ | 1.4 | mm |
| $k$ | 7 | mm |
| $r$ | 20 | mm |
| $\alpha$ | 0.15 | mm |
| $\eta$ | 0.5 | mm |
| $Z$ | 1 | mm |
| $z$ | 0.15, 0.3, 0.45, 0.6, 0.75, 0.9 | mm |
| $d$ | 0.5, 1, 1.5, 2, 2.5, 3 | mm |
| $A$ | -32.2315 | |
| $B$ | 124.4438 | |
| $C$ | 49.0965 | |

## V. GEOMETRIC DESIGN OF THE CONIC SURFACE AND IMPLEMENTATION IN SOLIDWORKS

The conceptual design in Fig. 4 generates a spiral motion at the tip point. However, this spiral motion is not necessarily the Archimedean spiral as desired for our application. With the Archimedean spiral the radius is directly proportional to the angle of rotation. In our design this means that the amount of the deviation of the tip point P from the center (z) should be proportional to the amount of the translation of the cam (d). This can be achieved by a proper design of the profile of the surface of the cone.

The solution is based on writing the geometric equations related to Fig. 4 and numerically solving to find the coefficients of a second order polynomial representing the curve of the cone. A straightforward application of geometric and trigonometric relations leads to the equations (1)-(5). We want to achieve the proportional relation in (6) by fitting a second order polynomial to the curve of the cone as in (7). The physical meanings of the parameters in (1)-(5) are revealed in Fig. 4.

$$e = \sqrt{(d+k)^2 + l^2} \quad (1)$$
$$\gamma = \sin^{-1}\frac{l}{e} \quad (2)$$
$$\theta = \sin^{-1}\frac{z}{r} \quad (3)$$
$$s = e\sin(\theta + \gamma) - l \quad (4)$$
$$f = \sqrt{e^2 - (l+s)^2} - k \quad (5)$$
$$z = \frac{\alpha}{\eta}d \quad (6)$$
$$f' = As^2 + Bs + C \quad (7)$$

In our mechanical design the lengths $l$, $k$, and $r$ are fixed. In (6) $\alpha$ is the distance between the scan lines which is desired to be fixed at 0.15 mm; $\eta$ is the distance of translation of the cam in one rotation and it is chosen as 0.5 mm. In our formulation $l$, $k$, $r$, $\alpha$, and $\eta$ are constants; $e$, $s$, $d$, $f$, $\gamma$, $\theta$, and $z$ are variables.

The mathematical problem is to minimize the error between the geometrically calculated $f$ in (5) and the curve to be fit represented by $f'$ in (7). This optimization is to be performed by modifying the parameters $A$, $B$, and $C$ in (7). The minimization should be performed in the range of $z$ that covers the desired region of scan, let's say in the range [0, Z].

The problem statement is

*Minimize*
$$E = (f' - f)^2 \quad (8)$$
*subject to*
(1)-(7) and $0 \le z \le Z$
*to determine*
$A, B, C.$

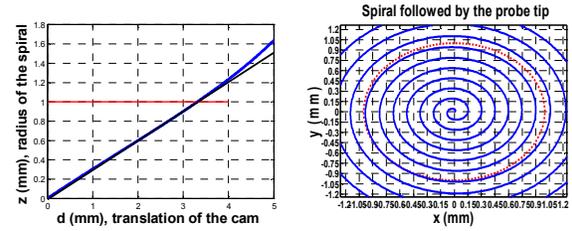

Fig. 5. Result of the optimization. Left: the relation between spiral radius and translation of the cam; right: the spiral obtained by rotating the cam.

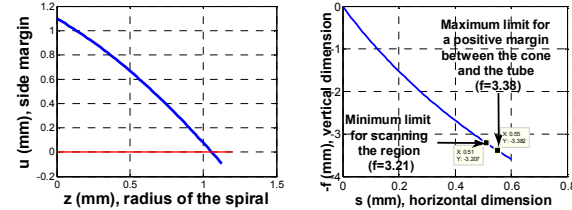

Fig. 6. Left: the margin between extremity of the conic structure and the inner surface of the tube; right: the resulting curve of the conic surface.

We performed optimization by using six values equally distributed in the range of interest $z \in [0,1]$ mm. We numerically solved this problem with the *fsolve()* function in the optimization toolbox of MATLAB. In Table I, we present the values of the constant parameters, the values of $z$ and $d$ used for optimization, and the resulting values of the parameters of the polynomial.

On the left of Fig. 5 we depict the relation between the spiral radius and the translation of the cam: linearity is achieved within the region of interest, $z \in [0,1]$ mm. On the right we show the resulting spiral for 8 rotations. The region of interest indicated by the dotted (red) circle is covered in less than 7 rotations. Within this region the radius is linearly proportional to the angle of rotation: an Archimedean spiral is achieved. Outside this region the linearity degrades.

We calculated the height (change in vertical distance) and inclination of the tip of the conic structure from the nominal position, with respect to the radius of the spiral. At 1 mm radius, the height of the tip is 0.025 mm; the inclination is 2.87 degrees. Both values remain in the acceptable range, less than 0.1 mm and less than 5 degrees, respectively. One should be reminded that the tissue remains in contact with the probe in these ranges: our system guarantees contact with the tissue throughout the scan.

A geometric constraint for the design is that the conic structure should not touch the inside surface of the covering tube. The inner diameter of the tube is 5 mm. The minimum margin between the conic structure and the outer tube occurs when the cam is translated to its maximum, to the outer edge of the conic surface, where $f=m$. At this point the margin, $u$, is given as in (9), where 2.5 is the inner radius of the outer tube.

$$u = 2.5 - (2(s+l)\cos\theta - l) \quad (9)$$

The left-hand side of Fig. 6 depicts this margin with respect to the spiral radius. It is observed that the margin remains positive when the spiral radius is 1 mm; therefore it is possible to fit such a conic inside a 5 mm tube and generate the spiral with it. The right-hand side of Fig. 6 shows the curve representing the surface of the conic structure. The arrows indicate the limit dimensions of *f* and *s* for scanning the region of interest ($z \in [0,1]$ mm) and for maintaining a positive margin between the conic structure and the outside tube. We choose to use the limit values for a positive margin, namely the maximum value of *f* is determined to be *m*=3.38 mm.

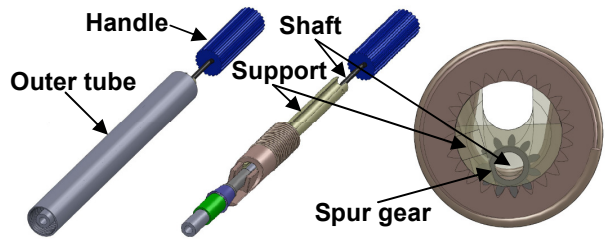

Fig. 7. The overall SolidWorks design of the scanning system. Left: with the outer tube; middle: without the outer tube; right: the support piece with the cam and spur-gear.

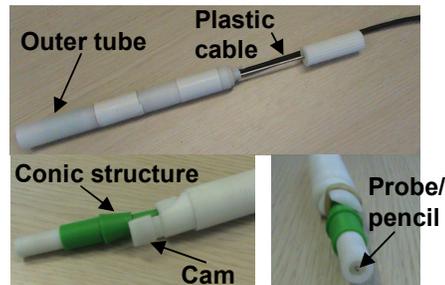

Fig. 8. Images of the 5:1 scale rapid-prototype of the scanning system.

We implemented the overall scanning system in SolidWorks. There are seven pieces in the overall mechanical design. These are the *conic structure*, the *cam*, an *outer tube*, a *support*, a *spur gear*, a *shaft*, and a *handle*. Fig. 7 shows the overall system with and without the outer tube. In this figure the support, shaft, and handle are visible. The handle is included only for manual drive of the system. As shown in the right-hand side of Fig. 7, the shaft is placed eccentric to the support. It rotates the spur gear located at the front part. This spur gear is coupled to the internal-spur-gear on the inner surface of the cam. In this way the rotation of the handle/motor is transferred to the cam. In Fig. 7 one can also see the eccentric canal to place the fiber cable along the support.

## VI. RAPID-PROTOTYPE AND MANUAL DRIVE

We implemented the system with rapid prototyping 5 times normal size. For rapid prototyping we used a Dimension 768 Series 3D Printer. The material used for building the prototype is ABS (acrylonitrile butadiene styrene) plastic. Fig. 8 shows the finished prototype. The top figure shows the overall system with the whole outer tube. The black cable replaces the fiber cable in the actual unity scale system. We used the outer plastic of an electric cable in this 5:1 scale prototype. In the bottom figures half of the outer tube is cut out. One can observe the cam and the conic structure. Since the plastic cable in this prototype is not stiff enough to force the conic structure to touch the cam, we included an extra elastic band in the design. This band ensures that the conic structure is always touching to the cam whatever the orientation is. The tip of the cam was modified to construct a bed for the elastic band.

In the 5:1 scale prototype we attached a pencil to the tip and drew spirals on paper by manually rotating the handle (Fig. 9). While drawing the spirals the tube was slightly pressed on the paper that was placed on a soft plastic bed. Once the handle was rotated from its initial to final position, the tube was raised away from the paper and the handle was rotated in the reverse direction. In this way the tip point was brought back to the center of the tube and the system was ready for another spiral drawing. On the right-hand side of Fig. 9, a sample spiral drawing is given. In

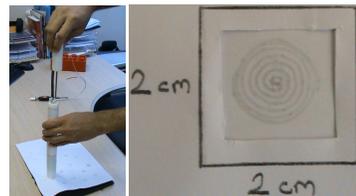

Fig. 9. Left: manual drive of the rapid-prototype with a pencil at the tip. Right: a sample spiral drawn by the tip point by manual drive.

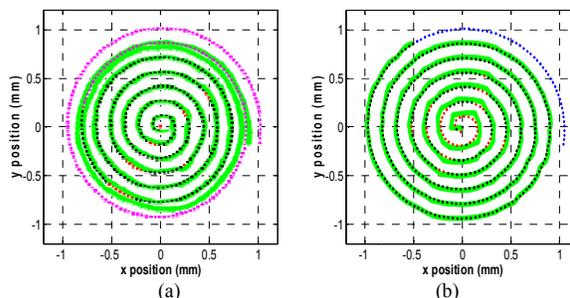

Fig. 10. (a) Solid (green) line: five times reduced figure of the spiral drawn by manual drive of the 5:1 scale prototype; dotted line: the ideal Archimedean spiral (black dots match with the drawn spiral; red dots mismatch; violet dots abruptly mismatch (irrelevant)). (b) Solid (green) line: the *probe trajectory* under the control of the Stäubli-Robot; dotted line: the ideal Archimedean spiral (*commanded trajectory*) commanded to the robot (black dots match with the *probe trajectory*; red dots mismatch; blue dots are irrelevant).

Fig. 10(a) we show a reproduction of this figure based on fivefold reduction and compare it with the desired Archimedean spiral in the unity scale dimensions. The ideal spiral is indicated by the dots in the figure. We would like to stress that the comparison in the following is based on the five times reduced image obtained with the 5:1 scale system. Therefore it cannot be considered as a validation of actual unity scale system; but it gives an impression of the performance of the 5:1 scale system.

The outer line of the spiral drawn in 5:1 scale corresponds to an approximate circle with 0.91 cm radius in unity scale. When we add 100 μm, which is the half of the minimum dimension of the field of view, the outcome is a circle with 1.01 mm radius and 3.2 mm$^2$ area. The distance between the center of lines is less than 0.18 mm (thickness of the green lines is approximately 0.03 mm). These results satisfy the requirements of minimum 3 mm$^2$ area and maximum 0.2 mm (200 μm) distance between scan lines.

As it is observed the drawn spiral almost perfectly follows the ideal spiral up to the last one-and-a-half times round. In this last part there is an abrupt mismatch indicated with light (violet) dots. The reasons for this abrupt mismatch are the pushing effect of the rubber band and the elasticity of the plastic cable. When the cam translates to its limit dimension, the rubber band applies pressure on the conic structure. This pressure makes the elastic cable extend. As a result the conic structure is also translated forward with the cam; therefore the radius of the spiral does not increase. Neither of these causes exists in the original design. In the original system the fiber cable is stiff enough to ensure that the conic structure touches the cam and to ensure that there is no extension. Therefore we can expect a better match to the Archimedean spiral in the outer regions with the actual unity scale system.

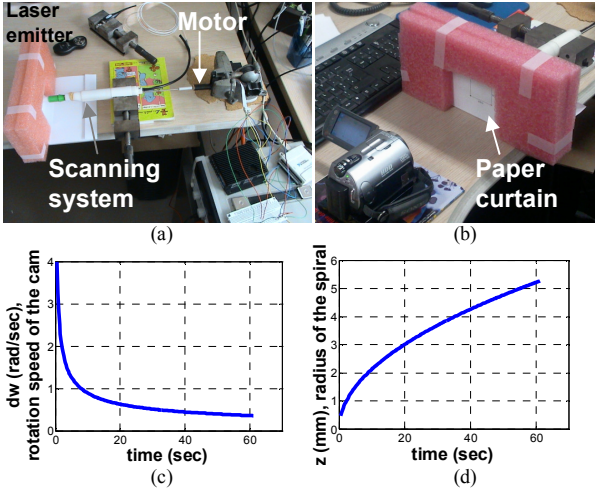

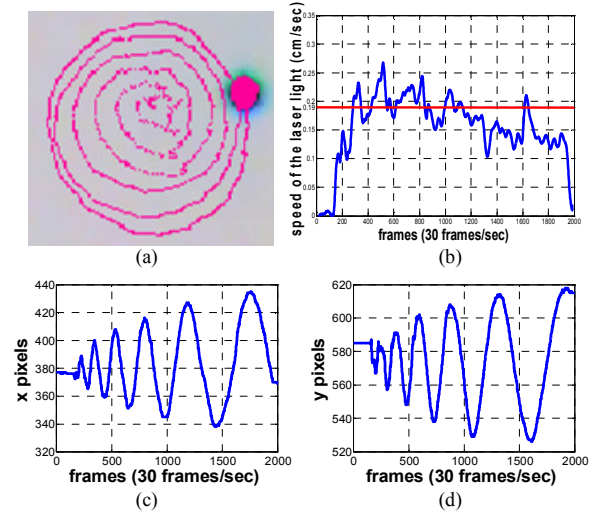

Fig. 11. (a) Experimental setup for the motor drive of the scanning system. (b) The movement of the laser light on the paper curtain is video-recorded. (c) The commanded speed of rotation for the cam. (d) The profile of the radius of the spiral throughout the motor drive.

Fig. 12. (a) The video-recorded trajectory of the laser light (b) The speed of the laser light. (c, d) The $x$ and $y$ positions measured in pixels.

We compare the spirals generated by the prototype and the Stäubli-Robot in order to quantify the degree of match with the ideal Archimedean spiral. In this comparison we ignore the abruptly mismatching last one-and-a-half times round indicated with the violet dots in Fig. 10(a).

In Fig. 10(b) we show the *probe trajectory* of the experiment presented in Fig. 3, with a green line of approximately 0.03 mm (30 μm) thickness. It should be noted that, unlike Fig. 10(a), the Fig. 10(b) is generated in actual dimensions depicted on the figure with the Stäubli-Robot. On this figure we again indicate the ideal Archimedean spiral, the *commanded trajectory*, with dots. The dots of the ideal spiral are black (dark) when their centers are on the green line and red (light in the inner part) when they are out. The outer part that is not traversed by the robot is irrelevant and colored in blue (light in the outer part). The same color measure applies to Fig. 10(a), except that the irrelevant part is in violet (light colored dots in the outer part).

Our measure is based on the ratio of the number of points outside the green line (red) to the total number of relevant points (red and black). This measure has a precision of approximately 15 μm, half of the thickness of the green line. The ratio is 0.89 with the Stäubli-Robot (50 red, 400 black) and 0.94 with our scanning system (20 red, 290 black). This means that in the 5:1 scale dimensions, our design can manage to follow the Archimedean spiral as well as the Stäubli-Robot does in unity scale. The Stäubli-Robot has difficulty following the ideal spiral in the very central part, where large acceleration is required.

## VII. EXPERIMENTAL RESULTS WITH MOTOR DRIVE

In this section we present the results obtained by driving the 5:1 scale prototype with a motor. Fig. 11(a) and (b) show the experimental setup with the scanning system, driving motor, motor controller, data acquisition box, laser light emitter, and the fiber optic cable passing through the scanning system. When the proximal end of the fiber cable is lighted with a laser beam the light reaches the distal end at the tip of the scanning system. The pink structure holds a paper curtain to visualize this laser light. The movement of the light is recorded by a camera at 30 Hz frame rate. We analyzed the video to identify the speed of the tip.

The requirement for the unity scale system is that the linear velocity of the probe should be less than 0.5 mm/sec. This corresponds to a speed limit of 2.5 mm/sec for our 5:1 scale prototype. We aim at a constant speed of 1.9 mm/sec, which corresponds to 1 minute duration to fulfill the scan. We calculate accordingly the rotational speed profile of the cam and the driving motor. The factor of reduction between the motor and the cam is 11:24. Fig. 11(c) shows the speed profile of the cam. The cam makes approximately 6.4 turns. As observed in Fig. 11(d) the radius of the spiral reaches 5 mm (1 mm in unity scale) in less than 60 seconds. After the completion of the spiral, the motor is rewound in less than 10 seconds, in order to bring the tip back to the center. This backward drive is free of imaging and is just for initialization of the system for a new scan.

Fig. 12(a) shows the spiral followed by the laser light. Like in the case of the manual drive, the diameter of the scanned region is approximately 9 mm, slightly less than the ideal (10 mm). The outermost line almost touches the previous. Again the pushing effect of the rubber band and extension of the cable are the causes of this discrepancy. The distances between the scan lines seem to be larger at the bottom side. This is because the paper curtain is not perfectly perpendicular to the surface scan of the probe.

The linear speed of the laser light is depicted in Fig. 12(b). The linear speed is around the expected, 1.9 mm/sec, but on average it is lower. Therefore the spiral ends before it reaches to the desired position: the cam makes slightly less than 6 turns instead of 6.4. This is because the cam started at a slightly backward position than where it should; therefore some time was spent before reaching the proper part of the conic surface. This is observed in the position profiles given in pixel numbers in Fig. 12(c) and (d) that the starting points are not at the center of the oscillating branches. The backward shift in starting position resulted in a decrease in the velocity profile.

The distance between the lines was expected to be 0.75 mm. In Fig. 12 it is larger in the middle phases, around 1 mm, and less in the last phase, around 0.25 mm. This variation is due to the rolling and stretching of the cable. With the plastic cable used in this prototype the conic structure rolled to some extent and this rolling pulled it upward, away from the scan surface. The upward movement generated a larger radius than expected. It should be noted that the plastic cable did not roll so much in the manual drive. This is because in the version of the manual drive the wires inside the cable, which is originally an electric cable, were kept and the overall cable was stiffer against the rolling motion. In the motor drive, we had to empty the wires inside in order to pass the fiber cable and this resulted in lower overall cable stiffness. In the original design there is no rolling and stretching, guaranteed by the stiff structure of the fiber cable.

In the latter stages the plastic rubber again pushed the conic structure forward. This pushing extended the plastic cable and the conic structure was translated downward towards the scan surface. Therefore, the radius was less than what it should be. A similar affect was observed also with the manual drive.

## VIII. Conclusion

In this paper we present a mechanical solution for scanning of soft tissues with a confocal microlaparoscope. The solution is based on using a conic structure and demonstrated with a 5:1 scale prototype.

The future implementation of the actual 1:1 scale system consists of producing the pieces in metal and assembling the overall system. Most of the parts, including the conic structure, are proper to be produced by conventional machining, which provides satisfactory precision. The remaining parts with rather complicated shapes, like the support and spur gear, can be produced by sintering technology. Sintering does not provide as good precision as conventional machining, therefore these pieces might need manual reworking by grinding and drilling. It should be noted that both conventional machining and sintering are low cost techniques frequently used in similar manufacturing. The cost of producing our pieces can be expected to be on the order of a few hundred Euros. The actual system will first be tested for precision and repeatability and be tested *in vivo* by a surgeon in an experimental pig operation.